# Publishing and linking transport data on the Web

Julien Plu
Université Montpellier 2
julien.plu@etud.univ-montp2.fr

François Scharffe
LIRMM
francois.scharffe@lirmm.fr

## ABSTRACT

Without Linked Data, transport data is limited to applications exclusively around transport. In this paper, we present a workflow for publishing and linking transport data on the Web. So we will be able to develop transport applications and to add other features which will be created from other datasets. This will be possible because transport data will be linked to these datasets.

We apply this workflow to two datasets: NEPTUNE, a French standard describing a transport line, and Passim, a directory containing relevant information on transport services, in every French city.

## Categories and Subject Descriptors

Open data, Semantic Web, transport, framework

## 1. INTRODUCTION

This project to build a framework for publishing and interlinking transport data on the Web was developed in collaboration between LIRMM[1], Université Montpellier 2[2] and CETE[3] Méditerranée. The feasibility of publishing and interconnecting transport transport into linked data on the web will be studied concretely from two data sources: the directory information services Passim and XML files corresponding to the NEPTUNE format describing public transport routes. Open Data, in France, has been booming for the last 2 years. Open Data is the publication of public data, free of charge and in open formats so that people who want to use them can do it. This phenomenon is closely linked with the term Linked Data. It refers to a set of best practices for publishing and interlinking structured data on the Web. These best practices were introduced by Tim Berners-Lee and have become known as the Linked Data principles. These principles are 1) use URIs as names for things, 2) use URIs which are dereferenceable, 3) use RDF and SPARQL standards, 4) link URIs between them. So if all data is linked, a huge graph is created, which forms the Web of data (or Semantic Web). The Semantic Web, term introduced by Tim Berners-Lee [1], aims at putting data on the Web in a form that machines can naturally understand, so web content can be treated directly or indirectly by machines. This is done with the help of ontologies. An ontology, according to Tom Gruber [2] is the specification of a conceptualization. A conceptualization is an abstract and simplified view of the world that we want to represent. More simply an ontology represents knowledge as a set of concepts within a domain, and the relationships between those concepts.

### 1.1 Passim

The directory Passim [3] is published by the CERTU[4]. It identifies and provides a list of information services on French passenger transport and other mobility services. Its content is managed by the CETE Méditerranée. In practice, the directory is a website which retrieves the relevant services for a city or territory in France, distinguishing between modes or types of transport (car, transit, etc.) and perimeter (urban, departmental, and regional). Services are at least web sites, sometimes including phone services, or mobile applications (in the future it could also be Web services). The directory contains freely reusable data (now published in the open data portal of the French government [4]). This directory could be extended for example with references to other datasets. Its format is published in CSV[5] and is represented as follows (translated):

Sheet number;Service Name;Coverage service;Region;Department;City;Modes of transport;Type of service;Network accessibility for disabled person;Land informations;Website;Website accessibility for disabled person;Information points ;Remark;Comments;Sms;Mobile application;List of cities covered (Postal code);Sheet created;Sheet modified

Each column is separated by the character ';'. The names of these columns are self-explanatory, here is a small example of a CSV line:

1;05voyageurs;départementale;Provence-Alpes-Côte d'Azur;Hautes-Alpes;N/A;Autocar, Covoiturage;Calcul d'itinéraire, Description du réseau,Horaires;Non;;http://www.05voyageurs.com;Non; ;;;;;09/06/2010;04/08/2011

A succession of ";" means that the column between the two ";" is empty.

### 1.2 NEPTUNE

NEPTUNE[6] is first the outcome of the European project TRIDENT and then of French work on the CHOUETTE application. NEPTUNE is a French Data Standard (NFP-99506) specifying the reference format for data exchange of theoretical transport offers, particularly useful for the development of multimodal information systems. NEPTUNE specifications consist in, on the one hand of a conceptual data model in UML (from the TRIDENT project, based on Transmodel V4.1[7])

---

concerning the definition of the network (lines, stops) and the theoretical service (races, schedules), and on the other hand of an XSD schema. The exchanged data is in the form of a directory with an XML file per line, each file describing all the information about a transport line (stops, schedules, etc.). NEPTUNE profile is fully compatible with the CHOUETTE software whose development is supported by the Ministry of Transport[8]. NEPTUNE will evolve in the context of works within the European standardization of the Netex[9] project. It should be noted that Google has specified an exchange text format, GTFS, which is the standard for publishing transport data into open data (mainly North American networks). The first NEPTUNE open data were published by the General Council of Gironde[10], others will follow (the General Council of Isère and the Saone et Loire). Here is an example of this format with the modeling of a bus stop:

```
<ChouettePTNetwork>

  <ChouetteLineDescription>

    <StopPoint>

      <objectId>NINOXE:StopPoint:15577811</objectId>

      <objectVersion>0</objectVersion>

                          <creationTime>2007-12-16T14:26:19.000+01:00</creationTime>

      <longitude>5.7949447631835940</longitude>

      <latitude>46.5263907175936000</latitude>

      <longLatType>WGS84</longLatType>

<containedIn>NINOXE:StopArea:1557779</containedIn>

      <name>Cimetière des Sauvages (A)</name>

    </StopPoint>

  </ChouetteLineDescription>

</ChouettePTNetwork>
```

Where :

- **ChouettePTNetwork**: a tag covering all information contained in the file;
- **ChouetteLineDescription**: a tag containing every descriptions a line may have;
- **StopPoint**: a tag describing the information a stop may have.

For more details on this format one could consult the standard online[11].

## 1.3 General publication process

The process of publishing and interlinking a dataset from a non-RDF format into RDF can be done in 4 steps:

- selection or creation of an ontology: this step aims at choosing an existing ontology describing the dataset to publish or eventually creating a new one;

- conversion to RDF: this step aims at converting the basic format of the dataset (XML, CSV, ...) in RDF, taking into account the chosen or created ontology;

- publication of RDF data: this step aims at making the data which have been previously converted into RDF available on the Web following the linked-data principles;

- interlinking with other datasets in RDF: this step aims at linking the RDF data with other datasets available elsewhere on the Web.

Here is a figure to illustrate these steps:

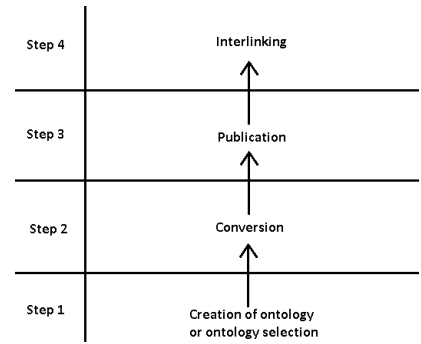

**Figure 1: diagram of the publication process**

These steps are more detailed in their respective sections. This process is taken from the process of the ANR project, DataLift[12].

## 1.4 Structure of the paper

In a first part we will present the two ontologies that were developed in order to describe data of the Passim directory, and NEPTUNE lines. In a second part we will see the means or methods used to convert these data into RDF. In a third part we will describe of the process of publishing the newly converted RDF data. Finally, in a fourth section we will detail the possible interlinking with other datasets.

## 2. RELATED WORK

In this section we will cover related work in transport data and formats, and data publication processes.

## 2.1 Open transport data

For the following reasons, transport data publication is popular in the UK:

- They are the pioneers of open data in Europe (culture of public transparency, Crown Copyright, etc.).

- The public transport sector has benefited from a decade of heavy investment in data standardization (Transport Direct project, standard NAPTAN, NPTG, TransXchange, etc.)[13].

A very good example of what can be done is posted on the United Kingdom open data website[14]. A lot of UK public transport data is on the Web, but little is available as linked data. Only a few datasets conform to Semantic Web standards have been identified: 1) the national repository of UK bus stops

---

(NAPTAN), 2) the Bus timetable of Manchester[15], 3) Maps of roads, railway stations and airport in the UK[16].

For a better understanding, a package containing all the linked data published by the Government of the United Kingdom is available online[17]. Despite the large number of U.S. transit systems whose tender is published in open data, their translation into linked data has only been the result of a few attempts and seems less advanced than in the United Kingdom.

Other data should follow, including all public transport timetables in the UK this year. There is also another ontology concerning transport, "The Tickets Ontology"[18].

In France, for the past two years, there is a strong interest for publishing public transport data as open data. For now, only three networks are online: Rennes, Nantes and Bordeaux. But with the ever-increasing number of data on the French open data portal there will certainly be more very quickly. From these data published in standard formats (NEPTUNE or GTFS), the technical work necessary to ensure their publication and their interlinking as linked data on the Web should not be too complex once the framework defined in this article will be completed.

## 2.2    Data publication process

### 2.2.1    Selection or creation of an ontology
The following tools are used for the selection of ontologies.

LOV[19] [5]: A RDF dataset identifying vocabularies that describe Semantic Web datasets, as well as the relationships between these vocabularies. Watson [6] and Falcons [7]: two ontologies search engines.

And the following one for the creation of ontologies.

Protégé[20]: an RDFS and OWL ontology editor. Neologism [8]: a Web application, derived from Drupal for publishing and creating ontologies. The ontologies Passim and NEPTUNE are published with this tool;

### 2.2.2    Conversion
Just like for the creation and selection of ontologies some tools are available on the Web.

CSV2RDF4LOD[21]: a tool to convert a CSV file in RDF. GRDDL [9]: a tool to convert a XML file in XSL. TriJplr[22]: a service generating RDF triples from URI. D2R Server [10]: a tool converting the contents of a relational database in RDF. D2R is also a web server for publishing the data.

### 2.2.3    Publication
We considered the following two tools for data publication, although other tools are available.

Virtuoso[23]: a tool to publish RDF data with many other possibilities (mapping databases, personal vocabulary, etc.).

OpenRDF Sesame [11]: a tool to publish RDF data, focusing primarily on the publication and database management functionalities. This list is not exhaustive. A benchmark study was conducted to compare all these tools [12].

### 2.2.4    Interlinking
The following tools were considered for interlinking datasets.

Silk [13]: provides a flexible declarative language to specify matching heuristics, which can combine different string comparators that can be digital as well as geographical;

LIMES [14]: implements a fast approach for the discovery of links at large scale, using metric spaces.

Recently another interlinking tool was developed within the project DataLift [15] along with a proposed framework for interlinking [16].

### 2.2.5    Existing frameworks
Currently there are already some frameworks to publish data in a raw format (CSV, XLS, XML, etc.) in RDF. But they have one (or more) gap(s) relative to what is presented in this article. The three most important are.

The Data Tank[24]: is a framework for publishing data in a RESTful way and easing the task of creating an application and allowing the publication and display of data in CSV, XLS and SHP (Shapefile). However it does not take the XML format, it does not publish data by using an ontology and does not makes interconnection. LDIF [17]: is a framework for retrieving all the data of interest from several sources and unifying them with a vocabulary by creating a custom URI and when we request this URI that affects multiple datasets simultaneously. LOD2 Stack[25]: is a suite of tool for data conversion, publication and interlinking. Its disadvantage is that it requires to use many tools requiring to learn many languages and specification formats.

## 3.    Data publication workflow

## 3.1    Selection or creation of an ontology
To take advantage of the semantics, we must model information, and for this, we rely on one or many ontologies. If the data is in a domain where there are already well-known ontologies, they are used. Otherwise, we have to create a new ontology, hoping it will be reused by others. Once the ontology created it must be published. Standards for writing ontologies are OWL [18] and RDF Schema [19]. RDF Schema provides basic elements for defining vocabularies or ontologies intended to structure RDF resources [20]: the main RDFS components are integrated into a more expressive (but more complex) ontology language, OWL.

### 3.1.1    Passim case
Concerning Passim there is no ontology that could have been specialized in this field: information transport. Therefore we have built the Passim ontology[26], linking it to the INSEE ontology[27] for the departments and regions, which originated in the work of DataLift[28]. The Passim ontology contains 4 classes and 18 properties. The list of classes being:

---

- TransportServiceInformation: this class represents a transport information service;

- Mode: this class represents the different modes of transport covered by the information service;

- Service: this class represents information services;

- Coverage: This class represents the geographic coverage of an information service.

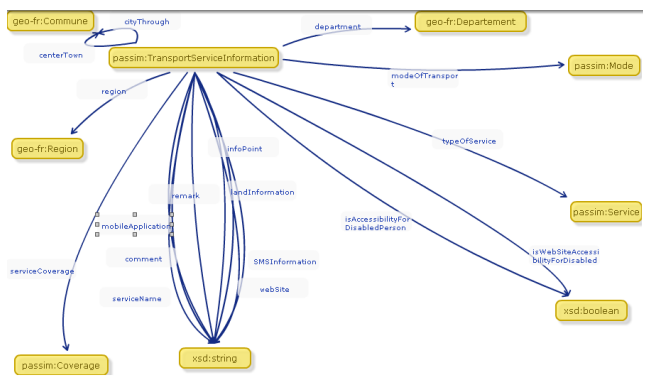

**Figure 2: diagram of the Passim ontology**

### 3.1.2   NEPTUNE case

As for NEPTUNE there are two potentially reusable ontologies:

- the one used by the site of the UK Government, NAPTAN[29];

- the one used by the city of Manchester, TRANSIT[30].

Although these English data formats are "cousins" of the NEPTUNE format, there will be undoubtedly links between vocabularies to do, there were not yet adapted to the NEPTUNE representation, so it was necessary, like for Passim, to develop an ontology for NEPTUNE[31]. Over time (from late 2012 at best), the European standard, Netex (under development) should unify the vocabulary and formats for data exchange in Europe for public transport data. At short term, we cannot link the NEPTUNE data with other Semantic Web data except via the geographic coordinates of stops; our vocabulary seems to diverge somewhat from existing ontologies despite a few similarities, like stops and roads. This ontology can therefore represent all data that may contain the original XML file (stop, route, schedule, etc.). It was particularly connected with the FOAF ontology and INSEE. The NEPTUNE ontology contains 50 classes and 125 properties.

## 3.2   Conversion to RDF

To make a particular dataset usable, we must publish in in a form allowing to know that the data is related to concepts defined in an ontology. A dataset is described as a list of facts represented as triples (subject predicate value). The Semantic Web standard for this is RDF.

### 3.2.1   Passim case

The exported CSV data from Passim website was converted to RDF with the tool provided by DataLift to convert the CSV into gross RDF without considering any ontology. Once the RDF file provided by the tool, the only thing left to do was to convert this

RDF file into another RDF file so that the data meets the Passim ontology. For this we must use the SPARQL [21] language and specifically the CONSTRUCT queries allowing to query RDF and producing RDF but modeled as desired, that is to say, in our case by taking into account the Passim ontology.

### 3.2.2   NEPTUNE case

Since NEPTUNE was an XML format we use another approach. We read and understood the corresponding standard and created a complex XSL transformation[32]. This sheet can then be run from the DataLift platform[33] to automate the conversion.

## 3.3   Publication of RDF data

Once in RDF format, we must publish these data (after ensuring that the ontologies which our data refers to are published on the Web). For this, we can:

- either simply publish the RDF file on the Web, i.e. make the file accessible from a browser by a URL;

- or publish them via a SPARQL endpoint. A SPARQL endpoint allows agents (human or machines) to query a knowledge base via the SPARQL query language.

The publication also includes the production of metadata allowing to list these data in various Semantic Web search engines, as Sindice [22]. For this there are certain prerogatives to be respected:

- using a vocabulary for describing data called VoID [23];

- creating a semantic sitemap [24], similar to a classic website sitemap, but adapted to the needs of Semantic Web search engines;

- creating a package on the Data Hub[34]. The Data Hub (formerly CKAN) references public datasets on the Web.

- manually referencing our dataset on Sindice or other search engines;

- provide URI dereferencing. A description of each resource will be provided if asked over the HTTP protocol;

- having a server to operate content negotiation (sending the corresponding page according to the agent of the request, HTML version if it is a human, or RDF version if it is a machine).

### 3.3.1   Passim and NEPTUNE case

Passim data when converted to RDF were published on a SPARQL endpoint[35], thus allowing to perform queries on this dataset such as the list of cities served by the company line TaM[36]:

```
SELECT DISTINCT ?city WHERE {
        ?s passim:serviceName ?o .
```

```
    ?s passim:cityThrough ?city .
    FILTER (?o = "TaM")
}
```

This dataset was also released on CKAN[37].

For NEPTUNE, only data from the Bordeaux city are available[38]. Also allowing to perform queries on the city of Bordeaux buses such as the names of all the bus stops in the city:

```
SELECT DISTINCT ?name WHERE {
    ?stop a neptune:StopPoint .
     ?stop neptune:name ?name .
}
```

## 3.4    Interlinking

In order to be connected to other Web datasets, the resources in the dataset need to be linked to equivalent resources in other datasets. Interlinking is used to retrieve all data from datasets which are connected to our.

### 3.4.1    Passim and NEPTUNE case

For Passim it is possible to link cities names, departments and regions with DBPedia [25] or the INSEE[39] or even Geonames[40]. DBPedia is the Semantic Web version of Wikipedia, more specifically this knowledge base contains all the information which are in the infoboxes in Wikipedia. Geonames is a large knowledge base containing information on cities, departments and French regions, but also other cities and countries worldwide.

## 4.    APPLICATION

With all these data we can imagine a useful application such as for example an application which, when you take a train or plane ticket for a city, lists all the activities that one can have in this city (visits, concerts (with information about singers or groups), shows, restaurants, ...), how to get there by public transports, the opportunity to share them on Twitter or Facebook, see their location with Google Earth, add the dates in your Google Calendar to remember what you have chosen to do and even know the current weather. Figure 3 details datasets used in this application.

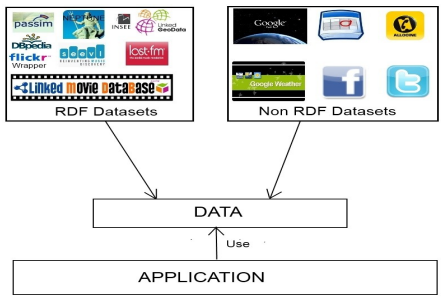

**Figure 3: datasets of this application**

The design and operation of the application will not be discussed, emphasis is mainly put on the use of these data and the connections between the RDF datasets. Here is how the datasets used are interconnected between themselves:

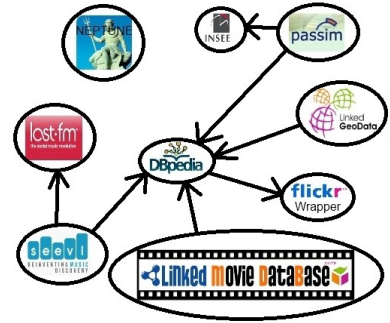

**Figure 4: Interconnection scheme between RDF data of the application**

As already explained NEPTUNE is not connected with other datasets. The big advantage is that when querying, for example, the Seevl Seevl, resources can be simultaneously retrieved from DBPedia and Last.fm datasets. This operation is more difficult to realize with datasets that are not in RDF. These datasets each have their usefulness:

- Seevl: information about musicians, bands, songs, etc. ;
- Last.fm: musical events, concerts, etc. ;
- DBPedia: large variety of domains ;
- Linked GeoData: RDF content of OpenStreetMap ;
- Flickr Wrapper: RDF wrapper of the Flickr API ;
- Linked Movie Database: information about actors, films, directors, etc. ;
- Google Earth: visualization of 3D buildings, satellite images, cities, streets, etc. ;
- Google Weather: know the weather at a specific location ;
- Google Calendar: calendar to add or view our appointments ;
- AlloCine: addresses of French cinema, timetable of films that are played.

## 5.    CONCLUSIONS

Public transport data can be used in countless professional and consumer applications as the data is published and reused, or as web services are available online. For example, a real-time map of all vehicles (trains, subways, buses, trams, taxis) of a city gives the user an overall view and possible delays: this application already exist (Swiss trains[41]), without needing Linked Data.

With this workflow we can develop applications using multiple datasets simultaneously. For example, it will be possible to use generic tools and well-formulated queries, to display restaurants and other activities available around each transit stop (with, if it's a cinema, movie times, if it is a restaurant, the menus and specialties, etc.), or to find tourist transportation routes based on destination and interests of the user.

We can also make small fun requests, for example, to know how many stops in France are called "Victor Hugo" or how many

---

stops are called after a French writer, and among these stops, which ones are located in a street that bears the same name.

# 6. ACKNOWLEDGMENTS

We would like to thank Patrick Gendre of CETE who is at the origin of this project.